\documentclass[twocolumn]{article}   
\usepackage[super,sort,comma]{natbib}
\usepackage{geometry}

\usepackage{fancyhdr}		

\usepackage[section]{placeins}   %

\usepackage{graphicx}

\makeatletter \renewcommand\@biblabel[1]{$^{#1}$} \makeatother
\newcommand{\cen}[1]{\begin{center} #1 \end{center}}

\usepackage[mathlines]{lineno}

\usepackage{hyperref}
\hypersetup{ colorlinks,
    citecolor=blue,
    filecolor=blue,
    linkcolor=blue,
    urlcolor=blue
}

\usepackage{xcolor}

\definecolor{gray}{rgb}{0.6,0.6,0.6}
\definecolor{red}{rgb}{0.85,0,0}
\definecolor{green}{rgb}{0,0.85,0}
\definecolor{blue}{rgb}{0,0,0.85}
\definecolor{beige}{rgb}{0.92,0.87,0.78}
\usepackage[all]{hypcap}    
\usepackage{amsmath}
\usepackage{amssymb}
\usepackage{booktabs}
\def \xbf{{\mathbf x}}

\def \Ibf{{\mathbf I}}
\def \vbf{{\mathbf v}}
\def \Wbf{{\mathbf W}}
\def \1bf{{\mathbf 1}}
\DeclareMathOperator*{\argmin}{arg\,min}

\begin{document}

\newgeometry{textwidth=14cm,textheight=18cm}

\onecolumn
{
\cen{\sf {\Large {\bfseries Extracting Tree-structures in CT data by Tracking Multiple Statistically Ranked Hypotheses } \\  
Raghavendra Selvan$^1$, Jens Petersen$^1$, Jesper H Pedersen$^2$, Marleen de Bruijne$^{1,3}$} \\
$^1 $Department of Computer Science, University of Copenhagen, Denmark \\
$^2 $Department of Thoracic Surgery, Rigshospitalet, University of Copenhagen, Denmark\\
$^3 $Biomedical Imaging Group Rotterdam, Department of Radiology and Nuclear Medicine, Erasmus MC, Rotterdam, The Netherlands

\today
}

\pagenumbering{roman}
\setcounter{page}{1}
\pagestyle{plain}
\begin{center}
Author email for correspondence: raghav@di.ku.dk   
\end{center}


\begin{abstract}
\noindent {\bf Purpose:} In this work, we adapt a method based on multiple hypothesis tracking (MHT) that has been shown to give state-of-the-art vessel segmentation results in interactive settings, for the
purpose of extracting trees. \\
{\bf Methods:} Regularly spaced tubular templates are fit to image data forming local hypotheses. These local hypotheses are then used to construct the MHT tree, which is then traversed to make segmentation decisions. Some critical parameters in the method, we base ours on, are scale-dependent and have an adverse effect when tracking structures of varying dimensions. We propose to use statistical ranking of local hypotheses in constructing the MHT tree which yields a probabilistic interpretation of scores across scales and helps alleviate the scale-dependence of MHT parameters. This enables our method to track trees starting from a single seed point. \\
{\bf Results:} The proposed method is evaluated on chest CT data to extract airway trees and coronary arteries and compared to relevant baselines. In both cases, we show that our method performs significantly better than the Original MHT method in semi-automatic setting. \\
{\bf Conclusions:} The statistical ranking of local hypotheses introduced allows the MHT method to be used in non-interactive settings yielding competitive results for segmenting tree-structures.  
\end{abstract}
}
\newpage
\restoregeometry
\twocolumn

\pagenumbering{arabic}
\setcounter{page}{1}
\pagestyle{fancy}

\section{Introduction}
\label{sec:intro}

In medical image analysis, reliable methods to extract airways, blood vessels and neuron tracks can have important clinical usage. For instance, airway tree extraction is used to study the morphology of airways, which is useful to derive biomarkers for diseases such as chronic obstructive pulmonary disease (COPD)~\cite{nakano2000computed,hasegawa2006airflow} and cystic fibrosis~\cite{kuo2017diagnosis}; and segmentation of coronary vessels is useful in prognosis of cardio-vascular diseases~\cite{rosamond2008heart}.

Several methods have been proposed to address tree segmentation tasks encountered in medical images. One common approach is to model appearance of local structures in the tree and extract the tree itself by formulating a global connectivity model.
Multi-scale vesselness filtering is a widely employed technique to enhance local tubular structures~\cite{frangi1998multiscale,bauer2008edge,yang2012automatic} like vessels and airways. Another approach is to detect local structures using multi-scale blob detectors and tracking these blobs to extract branches~\cite{selvan2017extraction}.
A comprehensive survey of vessel segmentation methods is presented in Lesage et al.~\cite{lesage2009review} and the class of methods we focus on in this contribution are addressed as template-matching based centerline tracking methods in this survey. A similar comparative study for airway segmentation methods was published in the EXACT'09 study~\cite{lo2012extraction}, wherein several methods were evaluated on chest CT data. One important take-away from this study was that there was scope for improvement in extracting missing airway branches. Another interesting point from the study was that more than half of the competing methods used some form of region growing in their segmentation procedure~\cite{feuerstein2009adaptive,bauer2009airway,wiemker2009simple,lee2009segmentation}. 

\begin{figure}[ht!]
	\centering
	\includegraphics{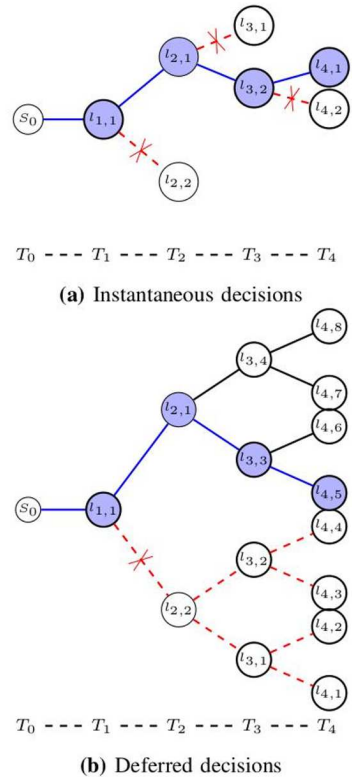}
	\caption{Instantaneous and deferred decisions illustrated using hypothesis trees. Hypothesis trees are shown for tracking steps $T_0,\dots,T_3$ with local hypothesis $i$ at step $t$ is denoted by $l_{t,i}$. In case of instantaneous decisions (left) only the best local hypothesis at each instant is retained (marked in blue). In the case of MHT with search depth $3$ (right), the decision at $T_1$ is based on the best global hypothesis at step $T_3$, marked with blue edges. Once the decision at $T_1$ has been made, hypotheses that are not children of the best node are discarded, shown in red.} 
\label{fig:comparison}
\end{figure}

In this work, we investigate the feasibility of using multiple hypothesis tracking (MHT)~\cite{reid1979algorithm,blackman2004multiple} -- a well known decision making paradigm in multi-target tracking -- for segmenting trees. In methods based on region growing~\cite{lo2010vessel} and conventional object tracking~\cite{yedidya2008tracking}, segmentation decisions are made instantaneously rendering them susceptibe to local anomalies due to pathologies, acquisition noise, interfering structures and other sources of noise. The primary contrasting feature of using MHT is that it is based on deferring decisions to a future step. From any given tracking step, all possible states corresponding to different solutions (local hypotheses) are maintained up to a predefined search depth (global hypotheses) in the form of a hypothesis tree. Decisions about which of the hypotheses to propagate and which ones to prune away at each step are made by tracing back the best hypothesis from the end of the hypothesis tree. Such deferred decisions, invariably, take more information into account and make the decisions at any given step more robust than instantaneous ones. The concept of instantaneous and deferred decisions are illustrated in Figure~\ref{fig:comparison}. 

In the work of Friman et al.~\cite{friman2010multiple}, MHT was used along with template matching for segmenting small and bright tubular structures from a dark background. This method when used in an interactive setting along with minimal paths has remained the best performing method in a coronary vessel segmentation challenge~\cite{schaap2009standardized}. However, to the best of our knowledge, it has not been applied to other tree extraction problems. Critical parameters used to maintain the MHT tree in this method are scale dependent; this has an adverse effect when tracking structures of varying dimensions, such as airway trees. We believe this method has potential to solve a wider range of problems if the limitations with scale dependence can be alleviated. 

We introduce statistical ranking of local hypotheses as a means to make local and global hypothesis scores independent of scales. This makes tuning of two crucial MHT parameters easier, allowing us to apply the method to extract trees and/or structures with branches of varying dimensions without user interaction. The work presented here is based on Friman et al.~\cite{friman2010multiple} and an extension of our previously published work~\cite{selvan2016extraction}. Compared to ~\cite{selvan2016extraction}, we present the proposed modifications formally and perform more comprehensive evaluation. We show airway extraction experiments on low-dose CT data from the Danish lung cancer screening trial~\cite{pedersen2009danish} and segmentation of coronary arteries from CT angiography data from the coronary challenge~\cite{schaap2009standardized}.  

\section{Method}\label{sec:meth}

In this section, we present an overview of the method in Friman et al.~\cite{friman2010multiple}, describe its limitations and present modifications that enable our proposed method to overcome these limitations. For convenience, the method in Friman et al.~\cite{friman2010multiple} will be addressed as the Original MHT method and ours, with the modifications, as the Modified MHT method.

\subsection{The Original MHT method~\cite{friman2010multiple}} 
\label{sec:orgMHT}

\begin{figure}[ht!]
	\centering
	\includegraphics{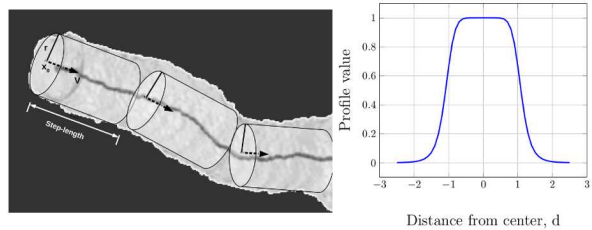}
	\caption{Illustrations of approximating an elongated structure of interest with a sequence of tubular segments (left) and a sample profile function used to construct tubular templates (right).}
	\label{fig:tubularModel}
\end{figure}

The primary objective formulated in the Original MHT method is to segment bright tubular structures from darker background based on their relative contrast. The bright, elongated structures of interest are modeled as sequences of tubular segments, as depicted in Figure~\ref{fig:tubularModel}. Segmentation is performed using a tracking approach, by analysing a multiple hypothesis tree comprising template-matched local hypotheses. The MHT method takes up a predict-update-aggregate procedure to build the hypothesis tree. 

{\bf Predict:} The prediction step is carried out from each tracking step where several local hypotheses are formulated at regular step lengths controlled by the \emph{step length factor} (used as a scaling of the local radius to obtain actual step length). The search span and number of local hypotheses at each step are controlled by the \emph{search angle} and \emph{number of angles} parameters, respectively. Local hypotheses are modeled using a parameterised template function which maps cross-sectional intensity variations of a tubular segment, of radius $r$ centered at $\xbf_0$ along the direction $\hat{\vbf}$, to a profile function with values between $0$ and $1$, given as, 
\begin{equation}
	T(\xbf;\xbf_0, \hat{\vbf}, r ) = \frac{r^{\gamma}}{(d_E^2(\xbf;\xbf_0,\hat{\vbf}))^{\gamma/2}+r^{\gamma}}. 
\label{eq:profile}
\end{equation}
Steepness of the profile function is controlled by $\gamma$ and $d_E^2(\xbf;\xbf_0,\hat{\vbf})$ is the squared Euclidean distance between any point $\xbf \in \mathbb{R}^3$ and axis of the tubular template along $\hat{\vbf}$. We retain $\gamma=8$ from Friman et al.~\cite{friman2010multiple} and the corresponding profile function is shown in Figure~\ref{fig:tubularModel} (right). 

{\bf Update:} The update step consists of performing the template-matching with the image data. This requires a model of the image neighbourhood, $\Ibf(\xbf)$ and is given as 
\begin{equation}
\Ibf(\xbf) = kT(\xbf;\xbf_0, \hat{\vbf}, r ) + m + \epsilon(\xbf),
\label{eq:image}
\end{equation}
where $k$ is the contrast and $m$ is the mean intensity and $\epsilon(\xbf)$ captures the noise due to modeling uncertainties, interfering structures, image artifacts or acquisition noise. Predicted local hypotheses are fit to the image data by solving the following weighted least-squares problem:
\begin{equation}
\argmin_{k,m,r,\xbf_0,\hat{\vbf}} || \Wbf(r,\xbf_0,\hat{\vbf})[kT(\xbf_0, \hat{\vbf}, r ) + m\1bf_n - \Ibf] || ^2,
\label{eq:min}
\end{equation}
where $\Wbf(r,\xbf_0,\hat{\vbf})$ is a weighting function with diagonal entries corresponding to an asymmetric Gaussian  centered at $\xbf_0$. It is used to localise the fitting procedure and $\Ibf$ is the image data with non-zero weights under this weighting function. The width of the weighting function is controlled by the \emph{weight window} parameter. 

The result of the template fitting procedure is an updated local hypothesis with estimates for the mean intensity and contrast values, which are then used to quantify the fitness of the local hypotheses.
A local hypothesis score is introduced to capture the fitness of local hypotheses: 
\begin{equation}
	l_i \triangleq \frac{k-m}{\text{std}(k)},
\label{eq:local}
\end{equation}
which can be seen as a measure of how different the bright tubular structure is from its background; this can be interpreted as the contrast signal-to-noise ratio (SNR), used to compare different local hypotheses. Local hypotheses below a threshold, referred to as the \emph{local threshold}, are discarded to control the number of candidate hypotheses. 

{\bf Aggregate:} An MHT tree of a predetermined search depth, $d$, is constructed using local hypotheses from each step as illustrated in Figure~\ref{fig:comparison}. A sequence of local hypotheses of length $d$ forms a global hypothesis with an average global score,
\begin{equation}
s_g = \frac{\sum_{i=1}^{d} l_{i}}{d}.
\label{eq:global}
\end{equation}
This score is compared with the \emph{global threshold} and all global hypotheses that do not exceed this threshold are discarded. Finally, the global hypothesis with the highest score is retained at the end of the tracking and output as the segmented result.

In summary, two categories of parameters are used to tune the Original MHT method. First category of parameters are related to generating local hypotheses: minimum and maximum radii, step length factor between successive tracking steps, window of the weighting function, maximum search angle and number of local hypotheses at each step. Second category of parameters are used to control the MHT behaviour: search depth, local and global thresholds. 

\subsection{Modifications to the Original MHT method}
\label{sec:ranking}

The Original MHT method was devised as an interactive method to segment bright, elongated structures. It is not immediately applicable for the automatic extraction of trees with branches of varying dimensions. In this section, some limitations of the Original MHT method in this regard are elaborated and modifications are proposed to overcome them. 

\subsubsection{Dealing with scale-dependence}

\begin{figure}[ht!]
\centering
	\includegraphics{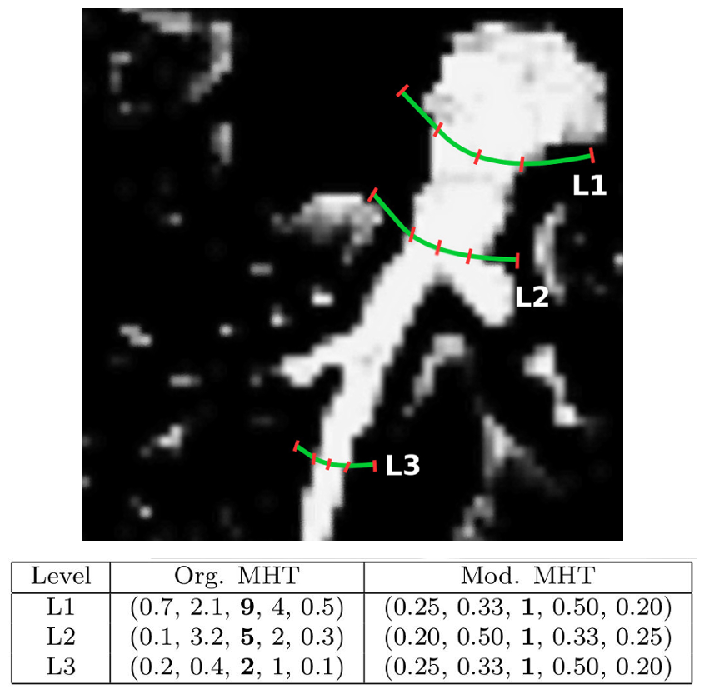}
\caption{Illustration of a branch of varying dimensions with three levels (L$_1$,L$_2$,L$_3$ highlighted in green); at each level five hypotheses (red marking) are tested. Local hypothesis scores for the original and Modified MHT methods are shown in tables on right. Notice how the scores of the best hypothesis at each scale varies for the Original MHT method.}
	\label{fig:varyingDim}
\end{figure}

An important factor to consider when tracking tree-structures with branches of varying dimensions is the range of their radius. With airway trees for instance, the radius of visible airways in CT typically ranges from $1$mm to $10$mm. When using the Original MHT method, such variations make it hard to obtain optimal parameter settings across scales. This behaviour is demonstrated for local hypothesis scores, computed using Eq.~\eqref{eq:local}, in Figure~\ref{fig:varyingDim}. Specifically, local hypothesis scores tend to decrease with scale, making it challenging to obtain a single optimal local threshold that works well across scales. When the local threshold is tuned to extract smaller branches, the chance of adding large and sub-optimal branches to the hypothesis tree is higher resulting in over-segmentation. Similarly, when the score is tuned for larger branches even the best candidate hypothesis for smaller branches might not qualify as valid candidates resulting in under-segmentation. 

To alleviate the aforementioned scale dependence, we devise a strategy of statistically ranking the local hypotheses based on their scores. That is, instead of using the scores in Eq.~\eqref{eq:local} directly, we sort them in decreasing order to derive a ranking based score at each step in the range $(0,1]$ . If $N$ local hypotheses are sampled at any given step, then the scores of the ranked local hypotheses are obtained as,
\begin{equation}
	l^{\prime}_i = \frac{1}{R_i} \quad \forall i \in 1\dots N,
	\label{eq:norm}
\end{equation}
where $R_i \in \{1,\dots,N\}$ is the rank of the $N$ local hypotheses. By this scheme of relative scoring of hypotheses, where the best local hypothesis is assigned $1$ at each tracking step, the problem of scale dependence of hypothesis scores is immediately alleviated. This is demonstrated in the table to the right in Figure~\ref{fig:varyingDim}. This procedure also provides a probabilistic interpretation of the hypothesis scores which is more useful in maintaining the MHT tree when local hypotheses are aggregated in Eq.~\eqref{eq:global}.

\subsubsection{Handling branching} Another factor to consider when tracking trees is bifurcations. In the Original MHT method, bifurcation detection is performed by clustering hypotheses based on their spatial location at every step using a spectral clustering algorithm~\cite{chung1997spectral}. However, there are no further details as to how the MHT tree is maintained after bifurcations. Once a bifurcation is detected, there are several ways to proceed. As the tracking happens per branch, new seed points can be added from the detected points of bifurcation and tracking can be restarted. This entails rebuilding of the MHT tree from each of the new seed points. The implementation of the Original MHT method, available as a MeVisLab\footnote{\url{http://mevislab.de/}} module, follows this strategy. This has a negative consequence of discarding the information aggregated until the step before branching. 

Another strategy that does not rely on rebuilding the MHT tree is to save the state of MHT tree at branching and resume tracking from each of the newly detected branching points separately. By resuming tracking from both branching points with the history of the parent branch we do not throw away information. In our Modified MHT method, this strategy is used to fully benefit from the MHT procedure.

\section{Experiments and Results}\label{sec:exp}

We evaluate the Modified MHT method for two applications: extraction of airway trees and segmentation of coronary arteries. We used the MeVisLab implementation of the Original MHT method for performing the Original MHT experiments and a customised module with our modifications on MeVisLab for the Modified MHT method. All experiments were performed on a laptop with $8$ cores and $32$ GB memory running Debian operating system. Details of preprocessing of data, experiments, error measures and results are presented next. 

\subsection{Airway tree extraction from chest CT}
\label{sec:aw}

{ Experiments were performed on a subset of 32 low-dose CT scans, from different subjects, chosen randomly from the Danish Lung Cancer Screening Trial~\cite{pedersen2009danish} dataset. Participants of the study were either current or former smokers between the age range 50 and 70 years, and had a history of smoking at least 20 pack years. The chosen subset of 32 scans has a distribution, that is similar to the DLCST study, with 19 COPD cases and 13 non-COPD cases. } 
The 3D images in this dataset have a slice spacing of 1 mm and in-plane resolution varying between $0.72$ to $0.78$ mm. The images were randomly split into training and test sets comprising of 16 images each. Performance of the method was compared with reference segmentations composed of union of two previous methods which were corrected by an expert user: the first method uses an appearance model based on a voxel classifier to distinguish airway voxels from background and uses region growing along with a vessel similarity measure to extract airways~\cite{lo2010vessel}, and the second method uses a similar voxel classifier but extracts airways by continually extending locally optimal paths~\cite{lo2009airway}. 
{The reference segmentations were constructed by the expert user -- a PhD student and one of the main organziers of the EXACT'09 Challenge~\cite{lo2012extraction} -- using a dedicated airway segmentation evaluation tool developed for the EXACT'09 Challenge. An airway branch was categorized as correct if there were no voxels beyond the airway wall, and wrong otherwise. Final corrected reference segmentations were obtained from the union of all branches classified as correct.} 

\subsubsection{Pre-processing of data}
\label{sec:probImages}

\begin{figure}[ht!]
\centering
	\includegraphics{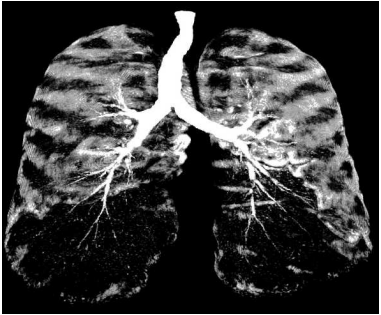}
\caption{Maximum intensity projection view of the probability image obtained from the voxel-classifier. Brighter regions correspond to higher probability, and hence more likely to belong to airways.}
\label{fig:prob}
\end{figure}

All CT images were pre-processed and converted into probability images using  a kNN-based voxel classifier trained to distinguish airway voxels from background~\cite{lo2010vessel}. Thus obtained probability images have probability close to $1$ in regions that are classified to be inside the airways and close to 0 outside. These images match the profile function described in Section~\ref{sec:orgMHT}, wherein, the structure of interest is bright (high probability) in a dark background. An example probability image is shown in Figure~\ref{fig:prob}. Noise in the image is due to several factors including acquisition, interfering vessels, ribs and lung tissue.

\subsubsection{Error measure}
\label{subsec:error}

{Performance of all four methods is evaluated based on accuracy of the extracted centerlines to give equal importance to large and small branches. This in in contrast to using voxel based measures such as dice overlap co-efficient which are biased towards large branches}. 
To standardise the evaluation procedure, centerlines were extracted from the binary segmentations from all methods using a 3D thinning algorithm~\cite{lee1994building}. 
 {The error measure is a symmetric distance between centerlines of the predicted and the reference segmentations}, given as,
\begin{align}
	d_{err} &= w\frac{\sum_{i=1}^{n_{op}} \min d_E(c_i,C_{ref})}{n_{op}} \nonumber \\
	&
	+ (1-w)\frac{\sum_{j=1}^{n_{ref}} \min d_E(c_j, C_{op})}{n_{ref}}.
\label{eq:error}
\end{align}
In the above equation, $C_{ref}$, $C_{op}$ are set of equidistant points on the centerlines of the reference and output segmentation results, respectively, comprising of $n_{ref}$ and $n_{op}$ number of points. $c_i, c_j \in \mathbb R^n$ are individual points on the centerlines, $d_E$ is the Euclidean distance and $w$ is a weight, such that $ 0 \leq w \leq 1$. Notice that the first term in Eq.~\eqref{eq:error} captures the distance between the two centerlines due to false positives, whereas the second term captures the distance due to false negatives. In this work we use $w=0.5$. Depending on the application the weight can be modified to obtain a desirable measure that reflects the sensitivity or specificity needs.

\subsubsection{Experiment set-up and parameter tuning}
\label{sec:awExp}

\begin{table}
		\caption{ Optimal parameters for both MHT methods (Org. and Mod.) based on training set for airway extraction along with the search range of parameters. Parameters with * were fixed based on the morphology of airways {on two scans from the training set to choose the largest feasible value}.
$^+$ We refer to the text in Section~\ref{sec:awExp} for explanation on \emph{Global threshold} for the Original MHT method and two different ranges for \emph{Weight window}.}
\label{tab:param}
\begin{center}
\scriptsize
	\begin{tabular}{ cccc } \toprule
		\textbf{Parameter/Method} & Search Range & Org. 	& Mod.	\\ \midrule
		Min. radius* (mm)	& 1 & 1		& 1		\\
		Max. radius* (mm)	& 10 & 10		& 10		\\
		Step length	& [1.0, 1.1, \dots, 2.0] & 1.5		& 1.1		\\
		Weight window$^+$	& [3, 4, 5] / [1, 2, \dots, 5]& 3.0 		& 1.0		\\ 
		Search depth	& 6 & 6		& 6		\\
		Search angle (deg.)&[30, 40, \dots, 70] & 60		& 70		\\
		Number of angles	& [1, 2, \dots, 5]& 3		& 2		\\
		Local thres.($T_\text{loc}$)	& [1.0, 2.0, \dots, 5.0]& 2.0 		& --		\\
		Global thres.$^+$ 	& $2T_\text{loc}$ / [0.5, 0.6, \dots, 0.9 ]& 4.0		& 0.7		\\ \bottomrule
    
    \end{tabular}
\end{center}
\end{table}

The Modified MHT method was compared to the Original MHT method and with region growing applied to both the probability and intensity images. Parameters of all the methods were tuned on the training set comprising 16 images and tested on an independent test set consisting of 16 images. Both the MHT methods have tunable parameters; the first set of parameters is related to the tubular template: minimum and maximum allowable radii of the templates, scaling factors for step length and weight window. A second category of parameters is related to multiple hypothesis tracking: search depth, search angle, number of angles, local and global hypothesis thresholds. Tuning both categories of parameters for the training set is cumbersome, so the minimum and maximum radii were fixed to $1$mm and $10$mm, respectively, based on prior knowledge about the morphology of airway trees. 
Increasing search depth can improve quality of segmentation decisions but it is constrained by the exponential increase in computation of additional hypotheses.
{With initial experiments, on two of the training scans, we tried out the largest search depth (6) for the chest CT data that resulted in a training time of about $8$ hours to try out the different parameter settings reported in Table~\ref{tab:param}.}
Only the remaining parameters were tuned to minimise the training error defined in Eq.~\eqref{eq:error} using grid search. The range of parameters searched over, and the optimal set of parameters obtained, for both MHT methods are summarised in Table~\ref{tab:param}. Due to the scale independence introduced by the statistical ranking of local hypotheses, the Modified MHT method does not have the local threshold as a parameter. The range of parameters for global threshold shown is for the Modified MHT method, as the global threshold for the Original MHT method was always set to be twice the optimal local threshold based on Friman et al.~\cite{friman2010multiple}. Both region growing methods (RG) only have the threshold parameter to be tuned. The optimal threshold for region growing on intensity of CT images was found to be $-995$ HU which is a low threshold, due to the leakage for three images in the training set for any higher thresholds. The optimal threshold for the case of probability images based on the training set performance was found to be $0.5$. All four methods require an initial seed point and the same seed point in the trachea, automatically extracted using the procedure described in Lo et al.~\cite{lo2010vessel}, was provided to all. 

We use the MHT module in MeVisLab provided by the authors in Friman et al.~\cite{friman2010multiple} to perform the Original MHT experiments. In this implementation, minimum value for the weight window factor is constrained to be $3$. However, for the Modified MHT method we tried a larger range $[1, 2, \dots, 5]$, and found the training error to be slightly lower (by $0.2$mm) when compared to using the smaller range $[3,4,5]$. 
We adhere to using this larger range for Modified MHT method, and a smaller range for Original MHT method,  also for the Coronary Artery Extraction experiments.

\subsubsection{Results}
\setcounter{footnote}{0}

\begin{table}[h]
	\caption{ Average error, $d_{err}$ in mm, for extraction of airways 
{reported for cases with COPD, No-COPD and combined}.}
	\label{tab:mht_results}
\label{tab:res}
\scriptsize
\begin{center}
    \begin{tabular}{ cccc}
    \toprule
    Method & COPD & No-COPD & All \\  \midrule
    RG (intensity) & $3.269 \pm 0.373 $ & $3.446 \pm 1.051 $ & $3.368 \pm 0.808 $ \\ 
    RG (probability) & $2.147 \pm 0.211 $ & $2.001 \pm 0.575 $ &  $2.064 \pm 0.447$  \\ 
    Original MHT & $ 3.770 \pm 0.771 $ & $ 4.109 \pm 1.756 $ &  $3.961 \pm 1.384$   \\ 
    \textbf{Modified MHT} & $1.916 \pm 0.426 $ & $ 1.940 \pm 0.567 $&  $1.929 \pm 0.494$   \\ \bottomrule
    \end{tabular}
\end{center}
\end{table}

\begin{figure}[h]
\centering
\includegraphics{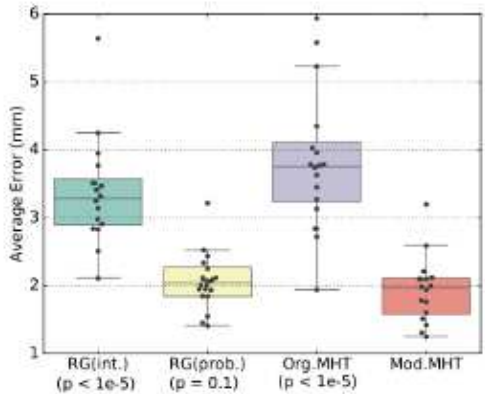}
		\caption{ Average error on 16 scans in the test set to compare the Modified MHT method with region growing (RG) on intensity images, region growing on probability images and the Original MHT method, visualised as standard box plots. Both MHT methods were applied to the probability images. 
{Pairwise significance test results comparing the performance of Modified MHT method with other methods based on Kruskal-Wallis tests are indicated in the bottom of the plot}.}
\label{fig:errPlot}
\end{figure}

\begin{figure}
\begin{center}
\includegraphics{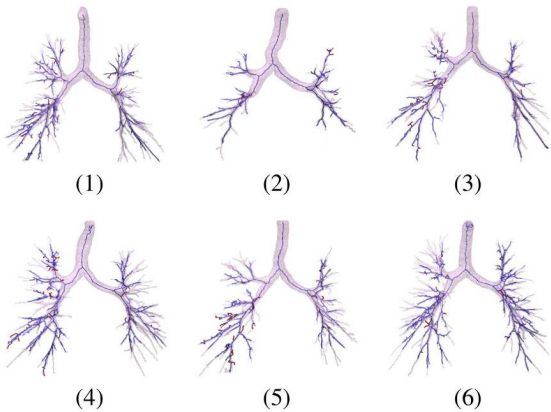}
\end{center}
		\caption{Six test case centerline results (thick lines) from the Modified MHT method are shown overlaid on the reference segmentations (background surface in pink). True positive portion of the extracted centerline is shown in blue and false positive portion in red. 
{ Counting trachea in the airway tree as generation $0$ and identifying subsequent generations at successive bifurcations, the modified MHT method is able to extract airways up to generation $8$ in most cases and even generation $11$ in some cases.}}
\label{fig:testPlots}
\end{figure}

Performance comparison was based on computing the average error measure in Eq.~\eqref{eq:error} and results for the test set are presented in Table~\ref{tab:mht_results} and visualised in Figure~\ref{fig:errPlot}. 
 {Differences between methods were tested for statistical significance using Kruskal-Wallis tests} %
. The Modified MHT method clearly shows an improvement when compared to the Original MHT method ($p < 1e-5$) and region growing on intensity images ($p < 1e-5$), but was not significantly better than region growing on probability images ($p = 0.1$). 
 {We also report the performance of each method for the COPD and non-COPD cases in Table~2 and find no significant difference in their performance between COPD and non-COPD cases}. Six of the test set results for the Modified MHT method are visualised in Figure~\ref{fig:testPlots} along with the reference segmentations. 

\subsection{Coronary Artery Extraction}
\label{sec:ves}
\footnotetext{Related to R\#1: Comment 2}

We next describe experiments on 3D CT angiography (CTA) images for segmenting coronary arteries evaluated on the Coronary Artery Challenge data~\footnote{http://coronary.bigr.nl/centerlines/index.php}. The challenge organisers allow methods to compete in three categories based on the extent of user interaction per vessel: automatic (no seed points), semi-automatic (one seed point) and interactive (more than one seed point). 
We evaluate the original and Modified MHT methods as stand-alone, semi-automatic methods and compare to the results of Friman et al.~\cite{friman2010multiple}. 

\subsubsection{Data and preprocessing}

The dataset consists of 32 CTA images, split into 8 images for training and the remaining 24 for testing. These 3D volumes have a mean voxel size of $0.32\times0.32\times~0.4$~mm$^3$. 
The objective of the challenge is to segment four vessels per dataset. To this end, four seed points per vessel are provided, of which one must be selected for semi-automatic methods:
\begin{itemize}
	\item Point S: start point of centerline 
	\item Point E: end point of centerline 
	\item Point B: a point about 3cm distal of the start point along the centerline
	\item Point A: a point inside distal part of the vessel.
\end{itemize}
In all images, three of the four vessels to be segmented were of the same type: right coronary artery (RCA), left anterior descending artery (LAD) and left circumflex artery (LCX). The fourth vessel, however, varied between images and was one of the large side branches (LSB). The reference standard for the training set was provided as manually drawn centerlines.

We closely adhere to the preprocessing performed in the Original MHT method~\cite{friman2008coronary,friman2010multiple}. Unsigned integers are used to represent the voxel intensities as gray value (GV), under a simple transformation of the corresponding Hounsfield Unit (HU): $HU(\mathbf{x}) = GV(\mathbf{x}) - 1024$. To enable improved vessel template matching, voxel intensity of the lung tissue  is raised to that of myocardial tissue and vessel calcifications are eliminated with the following thresholds:
\begin{equation}
	I(\mathbf{x}) = 
	\begin{cases} 
		t_{myo} & \text{if } I(\mathbf{x}) < t_{myo} \\ 
		I(\mathbf{x}) & \text{if } t_{myo} \leq I(\mathbf{x}) \leq t_{calc} 	 \\
		t_{myo} & \text{if } I(\mathbf{x}) > t_{calc},
	\end{cases}
\end{equation}
with $t_{myo} = 950$ and $t_{calc} = 1700$.

\subsubsection{Error measure}

We use two commonly reported measures on the Coronary Challenge website in our evaluations, that quantify completeness and accuracy of the segmented vessels, defined as OV (overlap) and AI (average distance inside vessel) in Schaap et al.~\cite{schaap2009standardized}, respectively. {These error measures are computed based on location of centerline points and the corresponding local radius information obtained from the segmentations.} 
{
\begin{itemize}
	\item {Overlap (OV):} It is similar to Dice overlap co-efficient and measures completeness of the tracked vessels. It is defined as OV in Schaap et.al~\cite{schaap2009standardized} as:
\begin{equation}
\text{OV} = \frac{|| \text{TPM}|| + || \text{TPR}||}{ || \text{TPM}|| + || \text{TPR}|| + || \text{FN}|| + || \text{FP}||}
\label{eq:ov}
\end{equation}
where points in the reference centerlines are marked TPR (true positive in reference) if the distance to at least one of the points on the output segmentation is less than the annotated radius, and FN (false negative) otherwise. Points of the output segmentation are marked TPM (true positive in method) if at least one point of the reference segmentation is at a distance less than the radius at that reference point, and FP (false positive) otherwise. $|| \cdot ||$ denotes the cardinality of the set of points. 
	\item {Accuracy (AI):} It is a measure of accuracy of the tracked centerline inside the vessel, obtained by computing the average minimum distance between each of the centerline points of the predicted and reference segmentations, if the distance is less than the annotated radius at the reference point. It is defined as AI in Schaap et.al~\cite{schaap2009standardized}.
\end{itemize}
}We also report overlap until first error (OF) and overlap with the clinically part of the vessel (OT) to enable better comparison with other methods in the challenge, computed similar to OV in Equation~\eqref{eq:ov}. {Further, the challenge organizers provide a score to measure agreement with observers for each of the measures which is also included in the evaluation. A score of 100 indicates that the result of the method is perfect, a score of 50 indicates performance similar to the observers and a score of 0 indicates complete failure.}

\subsubsection{Experiment set-up and parameter tuning}
\label{sec:corExp}

\begin{table}[h]
	\caption{Optimal parameters for both MHT methods after tuning on the training set for Coronary Artery Challenge to maximise overlap (OV) along with the search range of parameters. Parameters with * were fixed based on the morphology of vessels.
$^+$ We refer to the text in Section~\ref{sec:corExp} for explanation on \emph{Global threshold} for the Original MHT method and two different ranges for \emph{Weight window}.}

\label{tab:coronary}
\begin{center}
\scriptsize
	\begin{tabular}{ cccc } \toprule
		\textbf{Parameter/Method}& Search range &Org. 	& Mod.	\\ \midrule
		Min. radius* (mm) & 1 & 1 & 1 \\
		Max. radius* (mm) & 3 & 3 & 3 \\
		Search depth & 4 & 4 & 4 \\
		Weight Window$^+$	&[3, 4, 5] / [1, 2, \dots, 5] & 3 		& 1		\\ 
		Step Length	&[1.0, 1.1, \dots, 2.0]& 1.5		& 1.5		\\
		Search Angle (deg.)&[30, 40, \dots, 70]& 60		& 60		\\ 
		Number of angles	&[1, 2, \dots, 5]& 3		& 2		\\ 
		Local Thres. ($T_\text{loc}$)	& [1.0, 2.0, \dots, 5.0]  & 4.0 		& --		\\ 
		Global Thres.$^+$	& $2 T_\text{loc}$ / [0.55, 0.65,.., 0.95] & 8.0		& 0.95		\\ \bottomrule

    \end{tabular}
\end{center}
\end{table}

As with the airway tree extraction experiments in Section~\ref{sec:awExp}, most of the parameters are tuned for both MHT methods, while some are fixed based on prior knowledge. The minimum and maximum radii parameters are set to 1mm and 3mm, respectively. 
 { \emph{Search depth} was fixed to be $4$ which is the same as the one used in Original MHT work for fair comparison~\cite{friman2010multiple}.} 
Remaining parameters are tuned to maximise overlap, OV, on the training set and are summarised in Table~\ref{tab:coronary}. Further, as with the airway experiments, the starting value of weight window factor is $3$, and the global threshold is twice the local threshold, for the Original MHT method. Due to the varying quality of images across the training set, tracking from different seed points yields varying results. Based on the training starting tracking from Point B  and tracking in both directions turned out to be most useful for both MHT methods.
{ The final choice of seed point was based on estimating the overlap scores in the training set, starting from each of the four seed points and choosing the seed point which resulted in the highest overlap score (completeness).}

\subsubsection{Results}

\begin{table}
	\caption{ Accuracy (AI) and different overlap measures (OV, OF, OT) for the Original MHT method as interactive (int.) and semi-automatic (semi.) methods, and the Modified MHT method. 
{For each measure the corresponding agreement score with the observers is reported as Sc. columns}.} 
\label{tab:cor_results}
\scriptsize
\begin{center}\begin{tabular}{@{}lcccccccc@{}}
\toprule
Methods           & \multicolumn{2}{c}{AI} & \multicolumn{2}{c}{OV} & \multicolumn{2}{c}{OF} & \multicolumn{2}{c}{OT} \\ 
                  & (mm)       & Sc.      & (\%)       & Sc.      & (\%)       & Sc.      & (\%)       & Sc.      \\\midrule
MHT (int.)        &  0.230         &  47.9         &  98.521          &    84.0       &  67.612          &    72.8       &   91.783         &    84.5       \\
MHT (semi.)       &  0.388         &  29.8         &  83.078          &    40.6       &  44.078          &    31.7       &   70.992         &    44.1       \\
\textbf{Mod. MHT} &  0.368         &  29.3         &  98.681          &    63.6       &  70.034          &    53.9       &   93.372         &    67.7      \\ \bottomrule
\end{tabular}
\end{center}
\end{table}

\begin{figure}[!ht]
    \centering
        \includegraphics{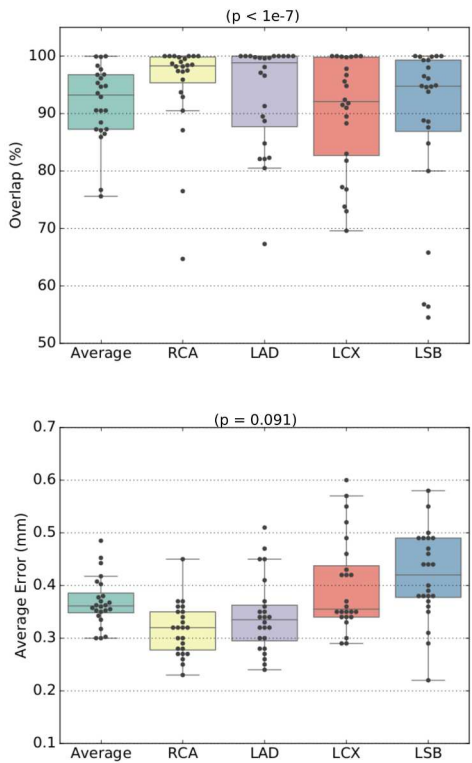}
		\caption{Overlap (OV) and accuracy (AI) measures for the Modified MHT method evaluated on the coronary challenge test set. 
{Significance results based on Kruskal-Wallis test comparing the Modified MHT method with the Original MHT method in semi-automatic setting is indicated on top of the figures.}}

\label{fig:coronary}
\end{figure}

We compare the Modified MHT method with the Original MHT method, evaluated as semi-automatic methods in the ongoing Coronary Artery Challenge evaluation. Results for these two submissions were processed by the challenge organisers. A version of the Original MHT method, used along with a minimal path algorithm and user interaction~\cite{friman2010multiple} has remained the best performing method in the challenge. 
The Modified MHT method scored $91.8\%$ overlap and accuracy of $0.38$mm, whereas the Original MHT method obtained $67.6\%$ overlap and accuracy of $0.39$mm. 
{There is significant improvement in overlap when compared to the Original MHT method $(p < 1e-7)$ as indicated by the Kruskal-Wallis test}. 
With these results our method stands as the third best performing method amongst all semi-automatic methods (requiring one seed point) reported on the challenge website.

Test set measures for each of the four vessels, along with the average score per dataset, are shown in Figure~\ref{fig:coronary} for the Modified MHT method. The first plot in Figure~\ref{fig:coronary} shows the overlap performance and it is clear that RCA, LAD and LCX vessels score high consistently and have few outliers. LSB, on the other hand, has four cases scoring less than $70\%$ in OV. One possible reason is that the vessel chosen as LSB was different across the dataset and the tuned parameters could not generalise to these four cases.
The second plot in Figure~\ref{fig:coronary} depicts the accuracy (AI) for the test set across the four vessels and again the average error in LSB is slightly higher than the rest. {The improvement in this measure is not significant when compared to the original MHT method $(p=0.091)$, indicating that the Original MHT method accurately segments incomplete vessels (OV $= 67.6\%$)}. In Table~\ref{tab:cor_results} we summarise the results for two versions of the Original MHT method and the Modified MHT method. 

\section{Discussion and Conclusions}\label{sec:disc}

In this work, we extended the template-matching based method for extraction of bright, elongated structures using multiple hypothesis tracking in Friman et al.~\cite{friman2010multiple}. We introduced statistical ranking of local hypotheses as a possible means to overcome scale dependence of important MHT parameters. 

A useful interpretation of the proposed ranking scheme is to see it as a measure of the relative significance of local hypotheses at each step. When aggregated over the depth of the hypothesis tree, it can be interpreted as the likelihood of the corresponding global hypothesis. Consider an example when the search depth is 10 and the global hypothesis threshold is $0.6$. For a global hypothesis (comprising of a sequence of 10 local hypotheses) to score above $0.6$ it would require its component local hypotheses to have been best locally at least in 6 of the 10 tracking steps, on average. This relative significance provides a probabilistic global hypothesis score, making it independent of the scale and variations of structures and removes the need to perform two level pruning of hypotheses -- using local and global hypothesis thresholds -- as done in the Original MHT method. Instead of explicitly pruning local hypotheses, we propagate all local hypotheses with their ranked scores from Eq.~\eqref{eq:norm} and rely on the deferred decisions of MHT algorithm to yield valid branches. 

One possible concern that arises when the scores in Eq.~\eqref{eq:local} are discarded for relative ranking in Eq.~\eqref{eq:norm} is the chance of poor candidate hypotheses performing well when ranked amongst equally poor counterparts. A strength of the deferred decision of MHT is that with adequate depth of the MHT tree such hypotheses are unlikely to be accepted, as they would have to perform well across multiple steps. Increasing search depth increases the computational expense as the number of hypotheses can grow exponentially. With both our experiments we observed that search depth of upto $6$ is feasible to run on limited computational resources, yielding promising results. Apart from occlusions due to noise in the data, the main factor contributing to missing branches in airways and vessels are small branches (as they are more susceptible to noise). 
{And in most cases, a hypothesis tree of search depth between $4-6$ can maintain adequate branch history allowing for a trade-off between under-segmentation of the structures and longer run time.}

Due to the proposed changes the Modified MHT method requires only one seed point per tree. In most cases, even the one required seed point can also be automatically provided, as in the case of airway experiments in Section~\ref{sec:aw}, and thus making the method fully automatic. We also presented an improvement to the bifurcation handling strategy by maintaining the history of the MHT tree across bifurcations instead of restarting tracking at bifurcation points.

We presented evaluations in Section~\ref{sec:exp}, where we first focused on extracting airway trees from chest CT data. To the extent of our knowledge, this is the first application of an MHT-based method to the task of airway extraction. The Modified MHT method was compared with region growing on intensity images, region growing on probability images and the Original MHT method. The Modified MHT method shows significant improvement in average centerline distance when compared to the Original MHT method $(p < 1e-5)$. The Modified MHT method required $15$ seconds on average, per CT scan, to extract airway trees starting from a single seed point. 
{The original MHT method when used in an automatic setting also uses $15$ seconds on average, whereas the region growing methods take less than $5$ seconds on average per scan.}

We also presented evaluations on the Coronary Challenge~\cite{schaap2009standardized} and compared the Modified MHT method with the Original MHT method used as a semi-automatic method. The interactive version of the Original MHT method, which is a combination of MHT, minimal paths and extensive user interaction (on average 1.5 clicks per vessel), is still the best performing method in the Coronary Artery Challenge. 
The Modified MHT method, however, shows significant improvement in the overlap measure (OV) when compared to the semi-automatic Original MHT method $(p<1e-7)$. The accuracy of the extracted centerlines (AI) from both semi-automatic versions of the MHT method are close, indicating that the Original MHT method (used semi-automatically) was accurate in the vessels that were segmented, but it under-segments the vessels indicated by the low OV score. This is likely due to dependence of the parameters on scale, which is overcome with the statistical ranking introduced in the Modified MHT method. In this work we evaluated the Modified MHT method using one seed point per vessel for the gains in some of the cases, while still competing in the semi-automatic category of the challenge. However, the Modified MHT method could also have been used with fewer than one seed point per vessel (one seed point per tree) and yielding similar performance, as we were able to extract substantial portions of the coronary tree starting from a single seed point. The computation time required for the Modified MHT method per vessel is $3$ seconds on average; thus not more than $15$ seconds per CTA image in the dataset. This makes it more appealing for clinical usage than the Original MHT method that similar computation time per time plus the time used for user interaction.

{The original MHT method used in an interactive setting takes up to 5 minutes per scan~\cite{friman2010multiple}.}

There are two recent semi-automatic methods based on diffusion tensor imaging techniques~\cite{cetin2013vessel,cetin2015higher} that perform better than the Modified MHT method in the OV measures as reported in the challenge website with a score of $97.3\%$ and $96.4\%$, respectively. However, in the OF measure these methods score $69.9\%$ and $77.4\%$, respectively, while the Modified MHT method obtained $77.9\%$. Also, the accuracy measure is comparable for all three methods ($\approx 0.35$ mm).
Three other methods in the fully automatic category have better OV measure by a small margin, between $1-2\%$, when compared to the Modified MHT method~\cite{yang2012automatic,zheng2013robust,bauer2008edge}. Again, the Modified MHT method obtains ~$3\%$ higher OF score when compared to these methods. One common feature across all these better performing methods is the pre-processing performed on the CT images to enhance vessel-like structures. For instance, in the currently best performing fully automatic method an improved version of Frangi's vesselness is used before centerline extraction~\cite{yang2012automatic}, and similarly in Bauer at al.~\cite{bauer2008edge} a tube detection filter is used and a pre-trained vessel detection filter is used in Zheng et al.~\cite{zheng2013robust}. In the Modified MHT method we use a simple template-matching step and, use of more sophisticated strategies as described in the other methods in the challenge should further improve the performance. For instance, we expect that the use of probability images obtained from a voxel classifier to distinguish vessel and non-vessel voxels instead of intensity images, as in the case of the airway experiments, could have been beneficial. 

In conclusion, we proposed modifications to the well established Original MHT method that yield significantly better results. Statistical ranking of local hypotheses yields a common interpretation of the global hypotheses of the MHT tree at all scales. 
{The improvements demonstrated with the modified MHT method in the reported measures translates into extraction of more complete airways and vessels, with fewer false positives. In the airway extraction experiments, the reduction in the average centerline distance has two contributing factors: fewer false positives and more complete airway trees. The latter of which -- detecting more complete airways in a COPD cohort -- can be of use clinically, as it is the detection of small airways that is harder and can be more useful in diagnosis of several pulmonary diseases. Further, the proposed modifications to the Original MHT method render it automatic and this can ease the clinical work-flow.}

\vspace*{-5mm}

\subsection*{Acknowledgements}
\vspace*{-2mm}

This work was funded by the Independent Research Fund Denmark (DFF) and Netherlands Organisation for Scientific Research (NWO). We thank the anonymous reviewers for their constructive feedback that has improved the manuscript substantially.

The authors have no conflicts to disclose.

\section*{References}
\addcontentsline{toc}{section}{\numberline{}References}

\newpage




\begin{thebibliography}{10}
\footnotesize
{
\bibitem{nakano2000computed}
Y.~Nakano et~al.,
\newblock Computed tomographic measurements of airway dimensions and emphysema
  in smokers: correlation with lung function,
\newblock American journal of respiratory and critical care medicine {\bf 162},
  1102--1108 (2000).

\bibitem{hasegawa2006airflow}
M.~Hasegawa, Y.~Nasuhara, Y.~Onodera, H.~Makita, K.~Nagai, S.~Fuke, Y.~Ito,
  T.~Betsuyaku, and M.~Nishimura,
\newblock Airflow limitation and airway dimensions in chronic obstructive
  pulmonary disease,
\newblock American journal of respiratory and critical care medicine {\bf 173}
  (2006).

\bibitem{kuo2017diagnosis}
W.~Kuo, M.~de~Bruijne, J.~Petersen, K.~Nasserinejad, H.~Ozturk, Y.~Chen,
  A.~Perez-Rovira, and H.~A. Tiddens,
\newblock Diagnosis of bronchiectasis and airway wall thickening in children
  with cystic fibrosis: {O}bjective airway-artery quantification,
\newblock European Radiology , 1--10 (2017).

\bibitem{rosamond2008heart}
W.~Rosamond et~al.,
\newblock Heart disease and stroke statistics—2008 update,
\newblock Circulation {\bf 117}, e25--e146 (2008).

\bibitem{frangi1998multiscale}
A.~F. Frangi, W.~J. Niessen, K.~L. Vincken, and M.~A. Viergever,
\newblock Multiscale vessel enhancement filtering,
\newblock in {\em International Conference on Medical Image Computing and
  Computer-Assisted Intervention}, pages 130--137, Springer, 1998.

\bibitem{bauer2008edge}
C.~Bauer and H.~Bischof,
\newblock Edge based tube detection for coronary artery centerline extraction,
\newblock The Insight Journal  (2008).

\bibitem{yang2012automatic}
G.~Yang, P.~Kitslaar, M.~Frenay, A.~Broersen, M.~J. Boogers, J.~J. Bax, J.~H.
  Reiber, and J.~Dijkstra,
\newblock Automatic centerline extraction of coronary arteries in coronary
  computed tomographic angiography,
\newblock The international journal of cardiovascular imaging {\bf 28} (2012).

\bibitem{selvan2017extraction}
R.~Selvan, J.~Petersen, J.~H. Pedersen, and M.~de~Bruijne,
\newblock Extraction of Airways with Probabilistic State-space Models and
  {B}ayesian Smoothing,
\newblock in {\em Graphs in Biomedical Image Analysis, Computational Anatomy
  and Imaging Genetics}, pages 53--63, Springer, 2017.

\bibitem{wink2004multiscale}
O.~Wink, W.~J. Niessen, and M.~A. Viergever,
\newblock Multiscale vessel tracking,
\newblock IEEE Transactions on Medical Imaging {\bf 23}, 130--133 (2004).

\bibitem{dijkstra1959note}
E.~W. Dijkstra,
\newblock A note on two problems in connexion with graphs,
\newblock Numerische mathematik {\bf 1}, 269--271 (1959).

\bibitem{benmansour2011tubular}
F.~Benmansour and L.~D. Cohen,
\newblock Tubular structure segmentation based on minimal path method and
  anisotropic enhancement,
\newblock International Journal of Computer Vision {\bf 92}, 192--210 (2011).

\bibitem{florin2005particle}
C.~Florin, N.~Paragios, and J.~Williams,
\newblock Particle filters, a quasi-monte carlo solution for segmentation of
  coronaries,
\newblock in {\em International Conference on Medical Image Computing and
  Computer-Assisted Intervention}, pages 246--253, Springer, 2005.

\bibitem{lesage2016adaptive}
D.~Lesage, E.~D. Angelini, G.~Funka-Lea, and I.~Bloch,
\newblock Adaptive particle filtering for coronary artery segmentation from 3D
  CT angiograms,
\newblock Computer Vision and Image Understanding {\bf 151}, 29--46 (2016).

\bibitem{lesage2009review}
D.~Lesage, E.~D. Angelini, I.~Bloch, and G.~Funka-Lea,
\newblock A review of {3D} vessel lumen segmentation techniques: {M}odels,
  features and extraction schemes,
\newblock Medical image analysis {\bf 13}, 819--845 (2009).

\bibitem{meng2017tracking}
Q.~Meng, H.~R. Roth, T.~Kitasaka, M.~Oda, J.~Ueno, and K.~Mori,
\newblock Tracking and Segmentation of the Airways in Chest {CT} Using a Fully
  Convolutional Network,
\newblock in {\em International Conference on Medical Image Computing and
  Computer-Assisted Intervention}, Springer, 2017.

\bibitem{lo2012extraction}
P.~Lo et~al.,
\newblock Extraction of airways from {CT (EXACT'09)},
\newblock IEEE Transactions on Medical Imaging {\bf 31}, 2093--2107 (2012).

\bibitem{feuerstein2009adaptive}
M.~Feuerstein, T.~Kitasaka, and K.~Mori,
\newblock Adaptive branch tracing and image sharpening for airway tree
  extraction in 3-D chest CT,
\newblock in {\em Proc. Second International Workshop on Pulmonary Image
  Analysis}, 2009.

\bibitem{bauer2009airway}
C.~Bauer, T.~Pock, H.~Bischof, and R.~Beichel,
\newblock Airway tree reconstruction based on tube detection,
\newblock in {\em Proc. of Second International Workshop on Pulmonary Image
  Analysis}, pages 203--213, 2009.

\bibitem{wiemker2009simple}
R.~Wiemker, T.~B{\"u}low, and C.~Lorenz,
\newblock A simple centricity-based region growing algorithm for the extraction
  of airways,
\newblock in {\em Proc. Second International Workshop on Pulmonary Image
  Analysis (MICCAI)}, pages 309--314, Citeseer, 2009.

\bibitem{lee2009segmentation}
J.~Lee and A.~P. Reeves,
\newblock Segmentation of the airway tree from chest {CT} using local volume of
  interest,
\newblock in {\em Proc. of Second International Workshop on Pulmonary Image
  Analysis}, pages 273--284, 2009.

\bibitem{reid1979algorithm}
D.~Reid,
\newblock An algorithm for tracking multiple targets,
\newblock IEEE transactions on Automatic Control {\bf 24}, 843--854 (1979).

\bibitem{blackman2004multiple}
S.~S. Blackman,
\newblock Multiple hypothesis tracking for multiple target tracking,
\newblock IEEE Aerospace and Electronic Systems Magazine {\bf 19}, 5--18
  (2004).

\bibitem{lo2010vessel}
P.~Lo, J.~Sporring, H.~Ashraf, J.~J. Pedersen, and M.~de~Bruijne,
\newblock Vessel-guided airway tree segmentation: {A} voxel classification
  approach,
\newblock Medical image analysis {\bf 14}, 527--538 (2010).

\bibitem{yedidya2008tracking}
T.~Yedidya and R.~Hartley,
\newblock Tracking of blood vessels in retinal images using {K}alman filter,
\newblock in {\em Computing: Techniques and Applications, 2008. DICTA'08.
  Digital Image}, pages 52--58, IEEE, 2008.

\bibitem{friman2010multiple}
O.~Friman, M.~Hindennach, C.~K{\"u}hnel, and H.-O. Peitgen,
\newblock Multiple hypothesis template tracking of small {3D} vessel
  structures,
\newblock Medical image analysis {\bf 14}, 160--171 (2010).

\bibitem{schaap2009standardized}
M.~Schaap et~al.,
\newblock Standardized evaluation methodology and reference database for
  evaluating coronary artery centerline extraction algorithms,
\newblock Medical image analysis {\bf 13}, 701--714 (2009).

\bibitem{selvan2016extraction}
R.~Selvan, J.~Petersen, J.~H. Pedersen, and M.~de~Bruijne,
\newblock Extraction of airway trees using multiple hypothesis tracking and
  template matching,
\newblock in {\em 6th International Workshop on Pulmonary Image Analysis},
  2016.

\bibitem{pedersen2009danish}
J.~H. Pedersen et~al.,
\newblock The {D}anish randomized lung cancer {CT} screening trial—overall
  design and results of the prevalence round,
\newblock Journal of Thoracic Oncology {\bf 4}, 608--614 (2009).

\bibitem{chung1997spectral}
F.~R. Chung,
\newblock {\em Spectral graph theory},
\newblock Number~92, American Mathematical Soc., 1997.

\bibitem{lo2009airway}
P.~Lo, J.~Sporring, J.~J.~H. Pedersen, and M.~de~Bruijne,
\newblock Airway tree extraction with locally optimal paths,
\newblock in {\em International Conference on Medical Image Computing and
  Computer-Assisted Intervention}, pages 51--58, Springer, 2009.

\bibitem{lee1994building}
T.-C. Lee, R.~L. Kashyap, and C.-N. Chu,
\newblock Building skeleton models via {3-D} medial surface axis thinning
  algorithms,
\newblock CVGIP: Graphical Models and Image Processing {\bf 56}, 462--478
  (1994).

\bibitem{friman2008coronary}
O.~Friman, C.~K{\"u}hnel, and H.-O. Peitgen,
\newblock Coronary centerline extraction using multiple hypothesis tracking and
  minimal paths,
\newblock in {\em International Conference on Medical Image Computing and
  Computer-Assisted Intervention}, volume~42, 2008.

\bibitem{cetin2013vessel}
S.~Cetin, A.~Demir, A.~Yezzi, M.~Degertekin, and G.~Unal,
\newblock Vessel tractography using an intensity based tensor model with branch
  detection,
\newblock IEEE transactions on medical imaging {\bf 32}, 348--363 (2013).

\bibitem{cetin2015higher}
S.~Cetin and G.~Unal,
\newblock A higher-order tensor vessel tractography for segmentation of
  vascular structures,
\newblock IEEE transactions on medical imaging {\bf 34}, 2172--2185 (2015).

\bibitem{zheng2013robust}
Y.~Zheng, H.~Tek, and G.~Funka-Lea,
\newblock Robust and accurate coronary artery centerline extraction in {CTA} by
  combining model-driven and data-driven approaches,
\newblock in {\em International Conference on Medical Image Computing and
  Computer-Assisted Intervention}, Springer, 2013.
}
\end{thebibliography}


\vspace*{-10mm}
\end{document}